\documentclass[3p, twocolumn, 12pt, preprint]{elsarticle}

\usepackage[utf8x]{inputenc}
\usepackage{subeqnarray}
\usepackage{amsfonts}
\usepackage{amsmath}
\usepackage{amssymb}
\usepackage{graphicx}
\usepackage{color}
\usepackage{booktabs}
\usepackage{multirow}
\usepackage{siunitx}

\usepackage{enumitem}
\usepackage{lineno,hyperref}
\modulolinenumbers[5]
\usepackage{natbib}
\usepackage{lipsum}
\makeatletter
\def\ps@pprintTitle{%
 \let\@oddhead\@empty
 \let\@evenhead\@empty
 \def\@oddfoot{}%
 \let\@evenfoot\@oddfoot}
\makeatother

\usepackage{float}
\usepackage[none]{hyphenat}

\begin{document}

\begin{frontmatter}

\title{An Interactive Segmentation Tool for Quantifying Fat in Lumbar Muscles using Axial Lumbar-Spine MRI}

\author[add1]{Joseph Antony}

\author[add1]{Kevin McGuinness}
\author[add2]{Neil Welch}
\author[add2]{Joe Coyle}
\author[add2]{Andy Franklyn-Miller}
\author[add1]{Noel E. O'Connor}
\author[add3]{Kieran Moran}

\address[add1]{ Insight Centre for Data Analytics, Dublin City University, Dublin, Ireland}
\address[add2]{Sports Medicine Department, Sports Surgery Clinic, Dublin, Ireland}
\address[add3]{School of Health and Human Performance, Dublin City University, Dublin, Ireland}



\begin{abstract}
In this paper we present an interactive tool that can be used to quantify fat infiltration in lumbar muscles, which is useful in studying fat infiltration and lower back pain (LBP) in adults. Currently, a qualitative assessment by visual grading via a 5-point scale is used to study fat infiltration in lumbar muscles from an axial view of lumbar-spine MR Images. However, a quantitative approach (on a continuous scale of 0-100\%) may provide a greater insight. In this paper, we propose a method to precisely quantify the fat deposition / infiltration in a user-defined region of the lumbar muscles, which may aid better diagnosis and analysis. The key steps are interactively segmenting the region of interest (ROI) from the lumbar muscles using the well known livewire technique, identifying fatty regions in the segmented region based on variable-selection of threshold and softness levels, automatically detecting the center of the spinal column and fragmenting the lumbar muscles into smaller regions with reference to the center of the spinal column, computing key parameters [such as total and region-wise fat content percentage, total-cross sectional area (TCSA) and functional cross-sectional area (FCSA)] and exporting the computations and associated patient information from the MRI, into a database. A standalone application using MATLAB R2014a was developed to perform the required computations along with an intuitive graphical user interface (GUI).
\end{abstract}


\end{frontmatter}

\section{Introduction}
It has been suggested that the fat infiltration in the lumbar multifidus and the lumbar erector spinae muscles are related to the muscle atrophy \cite{Kader2000} and consequently lower back pain \cite{Pezolato2012} in adults. One of the main reasons considered for such relationship is that the increased intramuscular fat deposits may affect the contractility of the muscles required for the control of spinal orientation and inter-vertebral motion \cite{Kjaer2007, Pezolato2012, Mengiardi2006, Kader2000}. However, the relationship between fat infiltration and lower back pain needs further investigation.  Critical to understanding such relationship is the accurate and precise quantification of fat infiltration in the lumbar muscles. 

To date, in the broader area of fat deposition in muscles per se, qualitative \cite{Kjaer2007, Dhooge2012, Demoulin2007, Kader2000, Barker2004, Flicker1993, Parkkola1993, Sorensen2006} as well as quantitative approaches \cite{Mengiardi2006, Elliott2005, Datin1996, Houchun2012, Bandekar2005, Bandekar2006} have been adopted. 

The most common approach by far is to use a qualitative evaluation of fat infiltration in lumbar multifidus muscles \cite{Kjaer2007, Kader2000}. Goutallier et. al. \cite{goutallier1994} have proposed a semi-quantitative method involving a scale of 0-4 to grade the fatty muscle degeneration in cuff ruptures using CT scan images.  Battaglia et. al. \cite{battaglia2014} have investigated and validated the reliability of Goutallier classification system (GCS) for grading fat content in the lumbar multifidus (LM) muscles using MRI. This involves the use of a visual grading system in a scale of 0-4 to categorize the fat deposition \cite{Mengiardi2006}, where grade 0 corresponds to ``No intramuscular fat", grade 1 corresponds to ``Some fatty streaks", grade 2 corresponds to ``Less fat than muscle", grade 3 corresponds to ``Equal fat and muscle" and grade 4 corresponds to ``more fat than muscle".  

The qualitative approach, using a visual grading system for studying fat depositions has a limitation. Minute changes in muscle composition and fat deposition may not be clearly visible, at times may be overlooked \cite{Mengiardi2006} and do not provide precision of measurement.  However, quantitative fat measurements provide useful information and interventions for investigators in preventive medicine, longitudinal studies and they are also useful to clinicians who study the implications of steatosis and pathophysiology of fat \cite{Houchun2012}.


We have adopted a quantitative approach to precisely quantify the amount of fat deposition in the lumbar muscles. The proposed method of quantifying fat infiltration in the lumbar muscles is integrated in an interactive tool as a supporting system to the physicians to make better diagnosis as well as to check the effectiveness of the exercises or workouts being prescribed \cite{Storheim2003, Hultman1993} in rehabilitation programs \cite{Mengiardi2006}. In addition we quantify the fat content in the erector spinae muscles in a region wise manner with respect to the centre of the spinal column \cite{Samagh2011, Richard1989, David2012}, which represents the axis of spinal rotation \cite{Samagh2011}. From a bio-mechanical perspective of lower back pain, damage to the muscle region further from the axis of spinal rotation may have greater effect on motor control and subsequent levels of pain \cite{Richard1989, David2012}, because the moment of force produced by the muscle is dependent not only on the amount of muscle or muscle force, but also the distribution of muscle relative to the axis of rotation (the moment of force $\tau = ||r||.||F||. sin (\theta)$, where $r$ is the (lever arm) displacement vector, $F$ is the force vector, $\theta$ is the angle between lever arm and force vector). 

There are five key steps in this process. The first step is defining the region of interest (ROI) \cite{Pezolato2012, Dhooge2012} in the MRI-defined lumbar muscles using the ``livewire" (intelligent scissors) interactive segmentation technique \cite{Mortensen1998}. The second step is detecting the fatty regions based on a threshold \cite{Ballerini2002, Ballerini2000} and softness level selected by the user, and computing the fat percentage \cite{Pezolato2012, Ballerini2002,Ballerini2000} as a result. The third step is automatically detecting the center of the spinal column. The fourth step is sub-dividing the ROI into smaller fragments with reference to the centre of the spinal column. The final step is computing the total cross-sectional area \cite{Pezolato2012, Dhooge2012, Elliott2008}, the functional cross-sectional area \cite{Pezolato2012, Dhooge2012, Ranson2006} and the fat content percentage in every region. A stand-alone graphical user interface (GUI) using Matlab R2010a was developed based on the five steps, with interactive controls for selecting ROI from the input image, threshold adjustment, softness level adjustment, displaying the intermediate results and appending the computed results into an existing database.

The main contributions of this work are automatically detecting the center of the spinal column, quantifying fat in the fragments of lumbar muscles with reference to the center of the spinal column, and development of a standalone application with intuitive graphical user interface (GUI). The key difference in our work with reference to earlier reported work \cite{Kjaer2007, Pezolato2012, Dhooge2012, Mengiardi2006, Kader2000, Ropponen2008} is the use of a sigmoid function for quantifying fat in the lumbar muscles, which provides an additional sharpness control along with the threshold for identifying the fatty regions in the lumbar muscles. 

Our previous work \cite{Antony2014} is extended in the following way: automatically detecting the center of spinal column to quantify fat in the fragments of lumbar muscles with reference to the center of spinal column, computing a global image threshold using Otsu's method \cite{Otsu1975} and using it as initial reference for identifying fatty regions in the region of interest and the use of livewire interactive segmentation \cite{Mortensen1998} for defining the region of interest. To automatically detect the spinal column two methods are proposed: 1) Using the spinal cord as reference, 2) Automatic region detection using HOG features and an SVM classifier.

\section{Materials and Methods}
\subsection{MRI data}
Image acquisition was performed in the Sports Surgery Clinic, Santry Demense, Santry, Dublin. Images were acquired on a General Electric (GE) Signa HDxt 3 tesla scanner using an 8 channel phased array spine coil. Axial T2 FRFSE (Fast Recovery Fast Spin Echo) sequences were acquired on patients under investigation for lower back pain. Imaging parameters included; 4000/108 Repetition time/Time to Echo (TR/TE), 320$\times$244 matrix, 20$\times$20 cm field of view, a slice thickness 4mm with a 1mm gap. A dataset consisting 156 MR lumbar spine images of 26 subjects under the study of lower back pain and fat infiltration in lumbar spine muscles was included in this study. In this paper, we have shown the analysis and results of two patients. Using a DICOM converter, the lumbar spine MR images were converted to PNG format for analysis in MATLAB.

\subsection{Interactive segmentation tool}
\begin{figure} [!ht]
\centering
\includegraphics[scale=0.25] {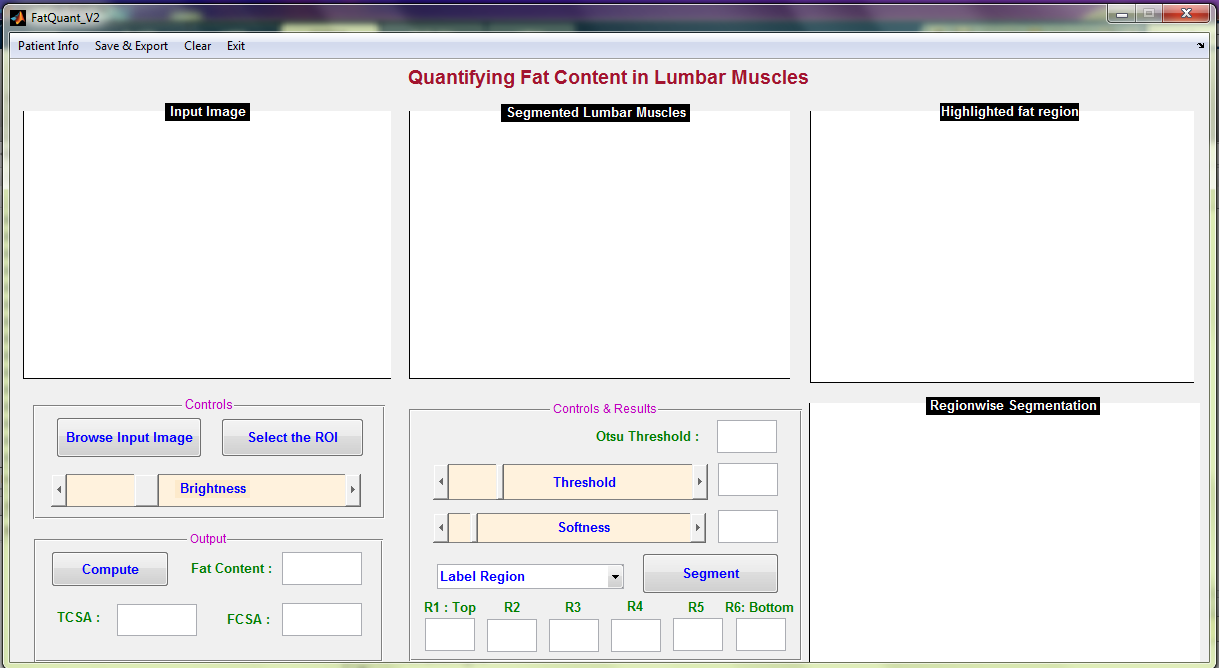}
\vspace{-5mm}
\caption{Screen shot of GUI}
\label{fig:GUI}
\end{figure}

A standalone Graphical User Interface (GUI) shown in Figure \ref{fig:GUI} using Matlab R2010a was developed with the essential interactive controls. Initially the GUI, allows the user to select an input image. Then the user can define the region of interest (ROI) by plotting a mask using the livewire (intelligent scissors) interactive segmentation technique \cite{Mortensen1998}. Once the mask is created interactively, the ROI is segmented and the grayscale image is displayed in the GUI. The global threshold to convert the grayscale image into a binary image is calculated using Otsu's method \cite{Otsu1975}. 
By default the threshold value is set at Otsu's threshold and the softness value is set at 0.2 (this value is based on empirical investigation). The softness value is mainly used for improving the visual clarity of fat regions by smoothing the edges of fat regions. For incremental variation of softness value in the steps of 0.1 from values 0 to 0.5, the fat percentage varies from 0.2 to 2 and the cross sectional area varies upto 3 $mm^2$. Initially, with pre-defined  threshold and softness, the fat regions are identified from the segmented lumbar muscle and displayed in the GUI. Based on visual inspection, suitable values for threshold and softness can be fixed by adjusting the `Threshold' and `Softness' sliding controls respectively.  The `Brightness' sliding control allows the user to adjust the brightness of the input image for better visualisation.

The total fat content percentage, total cross-sectional area (TCSA) and functional cross-sectional area (FCSA) in $mm^2$ can be calculated at any stage by pressing the `Compute' button. The computation of fat content percentage, TCSA, and FCSA is performed in line with the previous studies \cite{Dhooge2012, Pezolato2012, Mengiardi2006, Ranson2006}.

By using the drop down menu `Label Region' the user indicates the region of interest. The list of regions included in the menu are Right Erector Spinae muscles, Left Erector Spinae muscles, Left Lumbar Multifidus Muscles, Right Lumbar Multifidus Muscles, Right Psoas Muscles, and Left Psoas Muscles. 

The Erector Spinae (ES) Muscle is sub-divided into six fragments at equal intervals with reference to the center of spinal column and fat in each region is quantified. The center of spinal column is automatically detected for a given input image. The region wise quantification of fat  in the ES muscles, either on left or right side of the spinal column, is carried out by selecting `Segment' in the GUI. The ES muscle fragments are labelled R1 to R6 from top to bottom respectively, the fat percentage in each region is computed and displayed in the GUI. 
	
The GUI was iteratively developed based on feedback from experts.  The GUI includes the Otsu's threshold set as the initial reference, variable-selection of threshold and softness levels, computation of total and functional cross-sectional area, region-wise fragmentation of the ES muscle with reference to the centre of spinal column, which were based on experts opinion.

\vspace{-3mm}

\subsection{Defining the region of interest}
\begin{figure}[!ht]
\centering
\includegraphics[scale=0.7]{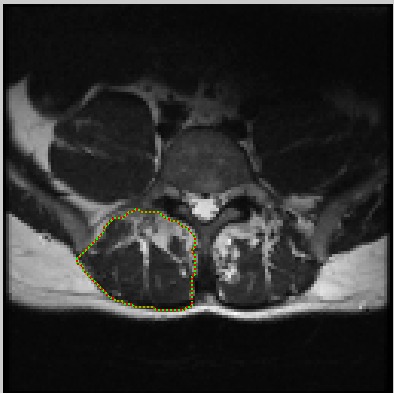}
\caption{MRI input image with user defined mask using Livewire interactive segmentation}
\vspace{-3mm}
\label{fig:Input image}
\end{figure}

The first step is selecting the region of interest (ROI) from the MRI-defined lumbar muscles, which can be any among the erector spinae (ES) muscles, lumbar multifidus muscles (LMM) or psaos muscles, located either on the right or the left side of the spinal column \cite{Pezolato2012, Dhooge2012}. The user has to define the ROI by plotting a mask over the input image  using livewire technique \cite{Mortensen1998}, as shown in Figure \ref{fig:Input image}. 

The livewire (or intelligent scissors) \cite{Mortensen1998} is a semi-automatic image segmentation technique that allows the user to interactively select the ROI on an input image using mouse clicks along the contour of the ROI. When the user starts the selection of the ROI with a mouse click, a virtual wire is created linking the first clicked point (referred to as an anchor) to the point where the mouse is over, following a path that is as close as possible to image features detected as edges using Dijkstra's lowest cost path algorithm. Figure \ref{fig:Input image} shows the result of a user segmentation using this tool.

The Livewire technique tends to work much slower in high resolution images, which would preclude its use. To resolve this, the input image was down sampled and the mask is defined in the low resolution image. 
The user defined mask is realized as a set of points in 2D coordinates, $f(x,y)$. Then using the set of points $f(x,y)$, the region inside the mask is cropped from the input image.

\vspace{-5mm}

\subsection{Identifying fat regions}
The pixel signal intensity (SI) variations between the muscle and the fat region can be used to distinguish the fat region from the muscle region \cite{Dhooge2012, Kader2000, Pezolato2012, Ranson2006, Datin1996}. By using an  appropriate threshold, the pixels in the fatty regions of the segmented lumbar muscles are detected \cite{Ballerini2002},\cite{Ballerini2000}. While the majority of the previous work \cite{Kjaer2007, Dhooge2012, Demoulin2007, Kader2000, Barker2004, Flicker1993, Parkkola1993} tends to use a hard threshold, the sigmoid function is proposed in this paper for setting the threshold level because it adds an additional softness level control for detecting the fatty regions. 

\subsubsection{Sigmoid function}
\begin{figure} [!ht]
\centering
\includegraphics[scale=0.5] {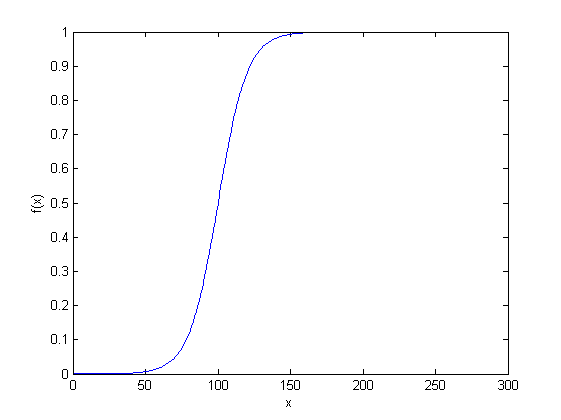}
\vspace{-3mm}
\caption{Plot of sigmoid function with a = 0.1 and c = 100}
\label{fig:sigmoid}
\end{figure}

The sigmoid function refers to a special case of a logistic function as shown in Figure \ref{fig:sigmoid}, defined by the equation:

\begin{equation}
s(x,a,c) = \frac {1}{1 + e ^ {- a (x - c)}} 
\end{equation}
where $c$ is the centre and $a$ is the slope control.

Every pixel $p(x,y)$ in the segmented lumbar muscle region is subjected to the sigmoid function $s(x,c,a)$ which gives a clear discrimination between the muscle region and the fatty region $p_{fat}(x,y)$ as shown in Fig. 4. 
\begin{equation}
p_{fat}(x,y) =
\begin{cases}
    1, & \text{if } p(x,y) \in \text{``fat" }
\\
    0, & \text{otherwise}
\end{cases}
\end{equation}

\begin{figure}[!ht]
\centering
\includegraphics[scale=0.25] {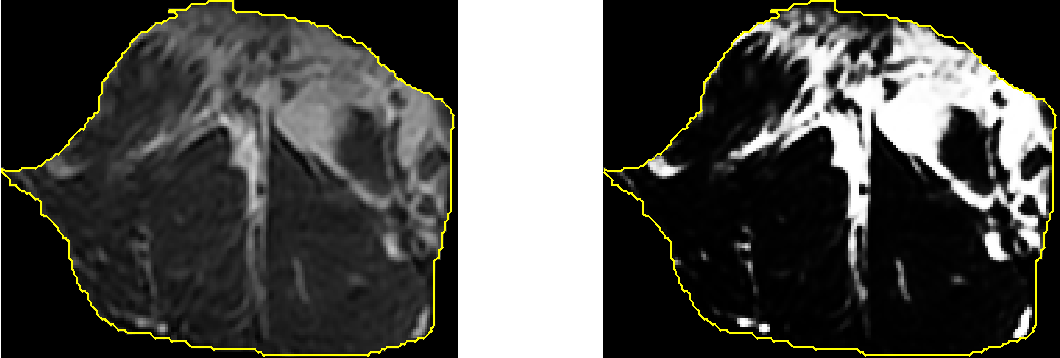}
\vspace{-3mm}
\caption{Detected fatty region from the segmented muscle region (with threshold 80 and softness 0.1)}
\label{fig:fatty region}
\end{figure}

In the sigmoid function, choosing different values for the centre $c$ is associated with the threshold selection for discerning the fatty region from the muscle region. Similarly varying the values of slope control $a$ in the sigmoid function is associated with the softness level of the discerned fatty region edges. Pixel $p(x,y)$ belongs to ``fat", if the pixel intensity is above the threshold selected by adjusting the centre $c$ value in the sigmoid function $s(x,c,a)$.

By calculating the ratio between the number of pixels $(N)$ in the segmented lumbar muscle region to the total pixels in the detected fatty region $\sum p_{fat}(x,y)$, the total fat content is calculated.
\begin{equation}
\text{Total fat content \%} = \left(\frac{\sum p_{fat}(x,y)}{N}\right) \text{ x } 100
\end{equation}
For example, considering the segmented region shown in Figure \ref{fig:fatty region}, the total pixels in the segmented lumbar muscle region were 21,156 and the pixels in the fatty region were 3,733 and the  computed total fat content was 17.6 \%.

\subsubsection{Thresholding (Otsu's method)}
\begin{figure}[!ht]
\centering
\includegraphics[scale=0.25] {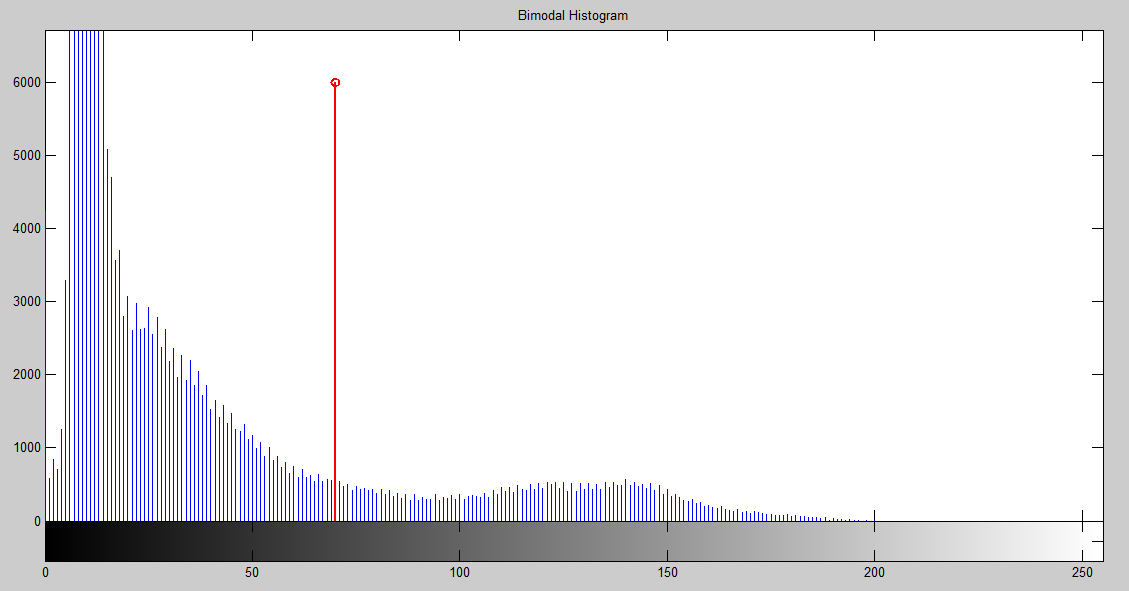}
\caption{Bimodal Histogram of an input image with Otsu's threshold}
\label{fig:Hist}
\end{figure}

The fat percentage in the ROI mainly depends on the choice of threshold value. To provide an initial reference to the user, the global image threshold calculated by Otsu's method is included in the GUI. 
Basically, Otsu's thresholding method considers that an image comprises of two classes of pixel intensity levels which can fall into a bi-modal histogram and an optimal threshold separating the two classes of pixels can be obtained such that their combined spread or intra-class variance is minimal. A bimodal histogram plot for an input image (Figure \ref{fig:Input image}) is shown in Figure \ref{fig:Hist} with the threshold level 70, calculated using Otsu's method. The whole input image was used to build the bimodal histogram. 

\vspace{-3mm}

\subsection{Automatic detection of the center of the spinal column}
The fat in the fragments of Erector Spinae (ES) muscles are quantified with reference to the center of spinal column. The center of spinal column can be selected by the user or it can be automatically detected. There are variations in the size and shape of the spinal column across different slices of MR Images of the same patient, which is the main challenge for automatic detection of the spinal column. We have adopted two different approaches for the automatic detection of the spinal column: a) with reference to spinal cord and b) using HOG features and an SVM classifier. 

\subsubsection{Detecting the spinal column with reference to the spinal cord}
The spinal column is not consistent in size and shape across different MR Images. In contrast, the spinal cord is relatively consistent in size, shape and intensity level across different MR images. 

\begin{figure}[!ht]
\centering
\includegraphics[scale=0.35] {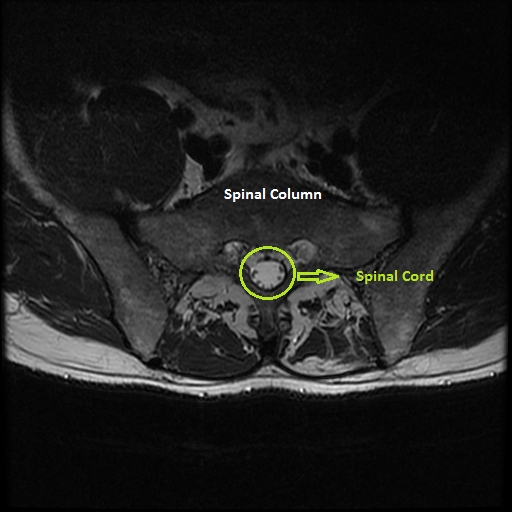}
\caption{The spinal column and spinal cord}
\label{fig:Spinalcord}
\end{figure}

The following steps are used to detect the center of  spinal column with reference to the center of spinal cord:

Step 1:  A central patch encompassing the spinal column and spinal cord is initially cropped from the input image. The central patch is used to avoid fat regions, while using intensity thresholds to detect the spinal cord, which is the brightest region in the cropped patch.

Step 2: Based on empirical investigation, an optimal threshold level (Otsu’s global threshold + 0.2) is used to detect the spinal cord over different MR Images. Using this threshold level the spinal cord is automatically detected in the process of grayscale to binary image conversion. 

Step 3: After detecting the spinal cord region, the centroid of the spinal cord is calculated.

Step 4: The center of the spinal column is approximately fixed 55 pixels above the center of spinal cord. This value was selected based on experiment results and after testing across different images.

The evaluation of this method was carried out by manually cropping the spinal column and calculating the centroids. The centroids obtained by the manual method were compared to centroids calculated from automatically detected spinal columns. This method is quick and precise, but for a small number of images there are slight variations (upto $\pm$ 7 pixels in X-coordinate and upto $\pm$ 15 pixels in Y-Coordinate) in the automatically detected center of the spinal column when compared to the actual center of spinal column. However, these slight variations do not affect the reference for region-wise fat quantification.

\subsubsection{Detecting the spinal column using  HOG features and SVM classifier}
The following steps are used to  detect the spinal column using an approach based on classifier:

Step 1: Initially, all the images are scaled to the same size $512\times512$. The images were split into training (75 \%) and test sets. 

Step 2: The image patches of size $ 50\times50$ comprising the spinal column are used as positive training samples. The image patches excluding fully/partially the spinal column regions are used as negative training samples. 

Step 3: As the training dataset was limited, to generate more positive samples, the image patches with spinal column were flipped in right to left orientation. 

Step 4: The Histogram of Oriented Gradients, a popular feature descriptor in computer vision is used to count the occurrences of gradient orientations in all the local patches of the images.
In the implementation, each cell size is
$2\times2$ pixels, and the orientation (0-180
degree) was separated into 9 histogram bins equally.
The $2\times2$ cells were combined in
to a block size $1\times1$ and histogram normalization was performed on the block. The descriptor is the vector of all components of the normalized cell responses from all of the blocks in the patch. Finally, a 5625-dimensional vector for a patch is extracted. The HOG features were extracted for positive and negative training patches and testing patches.

Step 5: The support vector machine (SVM) is a widely used classifier based on a supervised learning model in data analytics and pattern recognition. A linear SVM is trained with positive and negative training patches. In a 10-fold cross-validation among the training data, the classification accuracy was found to be 87.5 \% to 95 \%. After training, the SVM classifier is used to predict the unlabelled patches from test images. The  prediction accuracy is between 89 \% to 92 \%.

Step 6: In the process of automatically detecting the patch with the spinal column, a sliding-window technique was used to search exhaustively for the positive patch. A fixed size $50\times50$ window was used to scan the portions of the image surrounding the spinal column. For every patch detected by the window, the HOG features was extracted and tested with the learned classifier. Subsequently, a prediction score is assigned based on the SVM decision function. The patch with best score (minimal distance from the hyperplane) is selected as the outcome of successful detection. Finally, the centroid of the detected patch is calculated.

Though the initial experiments with this approach are encouraging, the accuracy of detection is found to be less when compared to the previous method i.e. using the spinal cord as the reference; this may be due to limited training images.

\subsection{Fragmenting ES muscles}
The next step is to quantify fat in fragmented regions of the erector spinae (ES) muscles. The segmented muscles could be sub-divided into many regions. We have subdivided the segmented muscles into six regions with reference to the centre of the spinal column, as shown in Figure \ref{fig:right seg} and Figure \ref{fig:left seg}. The use of six fragments was based solely on visual observation by a clinical biomechanist and clearly needs further research. Generally, the segmented muscle region is irregularly shaped. The boundary points are extracted and used to sub-divide the region into smaller segments. After obtaining the various segments, the fat content percentage in each segment is calculated.

The following steps are used to subdivide the segmented region and to perform the computations:

Step 1: The centre of the spinal column $c(x,y)$ is automatically detected for a given input image. 

\begin{figure}[!ht]
\centering
\includegraphics[scale=0.5] {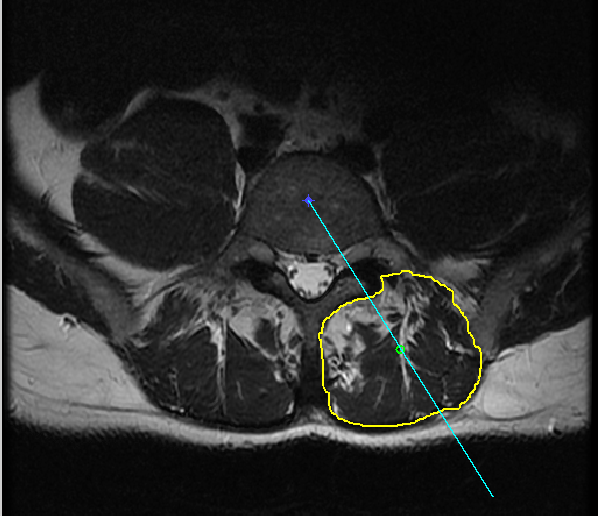}
\vspace{-3mm}
\caption{Input image with the radial line from the centre of spinal column (Right Erector Spinae Muscles).}
\label{fig: Right ES}
\end{figure}

\begin{figure}[!ht]
\centering
\includegraphics[scale=0.6] {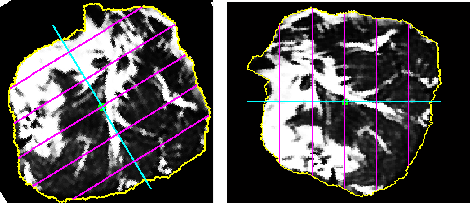}
\caption{Segmented Right Erector Spinae muscle rotated by angle $(\theta)$ in counter-clockwise direction.}
\label{fig:right seg}
\end{figure}

\begin{figure}[!ht]
\centering
\includegraphics[scale=0.6] {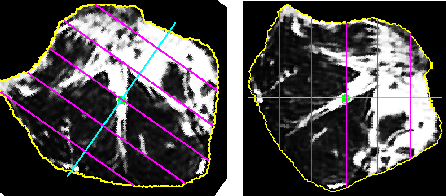}
\caption{Segmented Left Erector Spinae muscle rotated by angle$(180-\theta)$ in clockwise direction.}
\label{fig:left seg}
\end{figure}

Step 2: A radial line from the centre of the spinal column $c(x,y)$, which passes through the centroid of the segmented muscle region and that bisects the ES muscles is plotted, as shown in Figure \ref{fig: Right ES}.

Step 3:  Considering the radial line as vector $v1$ and a horizontal line through the centre of the spinal column $c(x,y)$ as vector $v2$, the angle $(\theta)$ between the vectors $v1$ and $v2$ is calculated.

Step 4:  The angle $(\theta)$ is used to identify, whether the segmented muscle region lies either on the right side or the left side of the spinal column.  If the angle $(\theta)$ is less than $90^o$ the segmented muscle region is considered to be on the right side and it is rotated by angle $(\theta)$ in the counter-clockwise direction as shown in Figure \ref{fig:right seg}, else the segmented muscle region is considered to be on the left side and it is rotated by an angle $(180-\theta)$ in the clockwise direction as shown in Figure \ref{fig:left seg}.

Step 5: From the segmented muscle region, the contour as a set of points, $f(x,y)$ is extracted. Subsequently the vectors $[X]$ and $[Y]$ pertaining to the contour points of the X-coordinates and Y-Coordinates, respectively, were extracted. 

Step 6:  The maximum and minimum values of $[X]$ and $[Y]$ are found, which gives the extremities of the irregular shaped segmented muscle region. 

Step 7: The length $(L)$ of the segmented region, which is the difference between the maxima and minima of $[X]$ is calculated. Further, the length $(L)$ is used to sub-divide the segmented muscle region.

Step 8: To have six sub-divisions, five equidistant vertical lines are plotted over the segmented lumbar muscle region at regular intervals $(L/6)$ from the minima of $[X]$. These vertical lines are plotted from minima of $[Y]$ to maxima of $[Y]$ so that every line touches the contours of the segmented muscle region as shown in Figure \ref{fig:right seg} and Figure \ref{fig:left seg}.
 
Step 9: Considering all seven hyperplanes, one each at the minima and maxima of $[X]$ and one at each of the five vertical lines, the fat content in six smaller segments are calculated.

Step 10: The fat content in every smaller segment is calculated by subjecting every pixel inside the region to the sigmoid function with the pre-selected threshold and softness level.  

\subsection{Computations}
 The physical pixel size $(psize)$ required for the calculation of TCSA and FCSA is read from MRI meta-data. The computations performed are:

\begin{equation}
\text{TCSA} = (N { \times } psize)
\end{equation}
\begin{equation}
\text{FCSA} = ((N - \sum p_{fat}(x,y) ){ \times } psize)
\end{equation}
where $N$ is the number of pixels in the segmented region, $\sum p_{fat}(x,y)$ is the total pixels in the fatty region. TCSA and FCSA are calculated in $mm^2$. Total fat content percentage is calculated as per equation (3).

\section{Results and Discussion}
\subsection{Quantifying fat in LM Muscles}
\begin{figure} [!ht]
\centering
\includegraphics[scale=0.7]{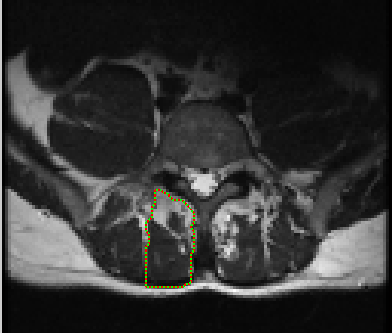}
\caption{Input image with ROI as LMM (left)}
\label{Reslmmleft}
\end{figure}

\begin{figure} [!ht]
\centering
\includegraphics[scale=0.7]{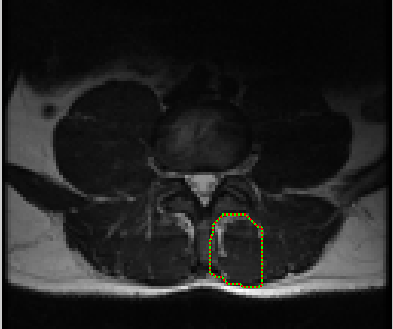}
\caption{Input image with ROI as LMM (Right)}
\label{Reslmmright}
\end{figure}

For the input images, with the defined region of interest (ROI)  being the lumbar multifidus muscles (LMM); the fat percentage, total cross sectional area (TCSA) and functional cross sectional area (FCSA) were calculated with Otus's threshold 70 (Figure \ref{Reslmmleft}), 56  (Figure \ref{Reslmmright}) and both with a softness level 0.2 shown in Table 1 and 2. Interestingly, both the images (Figure \ref{Reslmmleft} and Figure \ref{Reslmmright})would be classified in the same scale of fat infiltration (Grade 1 which corresponds to ``Some fatty streaks") in a 5-point grading system, though they clearly differ by 8.3\% of fat. Such quantification is useful for studies relating fat infiltrations in lumbar multifidus muscles to  lower back pain in adults because they offer a greater level of precision. 

\begin{table*}[!ht]
\centering
\caption{Fat Quantification: Lumbar Multifidus Muscles (left)}
\begin{tabular}{c c c c }
\hline
Segmented ROI & Fat (\%) & TCSA ($mm^2$) & FCSA ($mm^2$) \\ 
\hline
\begin{minipage}{0.08\textwidth}
\includegraphics[scale=0.4]{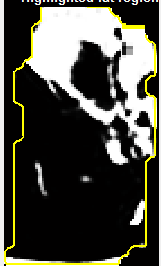}
\end{minipage}
& 25.8
& 33 
& 24 \\ 
\hline
\end{tabular}
\end{table*}

\begin{table*}[!ht]
\centering
\caption{Fat Quantification: Lumbar Multifidus Muscles (Right)}
\begin{tabular}{c c c c }
\hline
Segmented ROI & Fat (\%) & TCSA ($mm^2$) & FCSA ($mm^2$)   \\ 
\hline
\begin{minipage}{0.08\textwidth}
\includegraphics[scale=0.4]{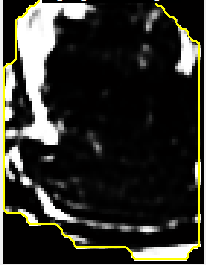}
\end{minipage}
& 17.4
& 25
& 21\\ 
\hline
\end{tabular}
\end{table*}

\subsection{Regionwise fat quantification in ES Muscles}
From the MR images of two different patients, the selected region of interest (ROI) being erector spinae Muscles (ES) either on left or right side of the spinal column the parameters (such as region wise fat content, total fat content, total cross sectional area (TCSA) and functional cross sectional area (FCSA)) were calculated. The results for Patient I and Patient II are shown in Table 3 and 4, respectively.

\begin{table*}[!ht]
\centering
\caption{Region-wise fat quantification results in ES Muscles - Patient I}
\begin{tabular}{ l  p{9cm} }
\hline
Segmented ES Muscles & Computations \\ 
\hline
\raisebox{-\totalheight}{\includegraphics[scale=0.9]{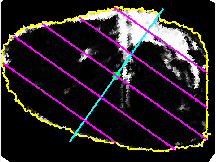}}
& 
\begin{itemize} [topsep = 0pt]
\item[] Region-wise Fat
  \begin{itemize}
  \item[] R1: 5.1 \% (Top), R2: 3.8 \%, R3: 1.4 \%,
  \item[] R4: 0.9 \%, R5: 0.6 \%, R6: 0.4 \% 
  \end{itemize}
\item[] Total Fat : 12.1 \%
\item[] TCSA : 44 $mm^2$, FCSA : 39 $mm^2$
\end{itemize} \\
\hline

\raisebox{-\totalheight}
{\includegraphics[scale=0.9]{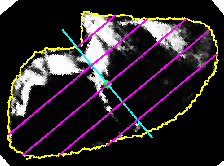}}
& 
\begin{itemize} [topsep = 0pt]
\item[] Region-wise Fat
  \begin{itemize}
  \item[] R1: 3.5 \% (Top), R2: 5.1 \%, R3: 2.9 \%,
  \item[] R4: 1.8 \%, R5: 0.8 \%, R6: 0.4 \% 
  \end{itemize}
\item[] Total Fat : 14.4 \%
\item[] TCSA : 45 $mm^2$, FCSA : 38 $mm^2$
\end{itemize} \\
\hline

\raisebox{-\totalheight}
{\includegraphics[scale=0.9]{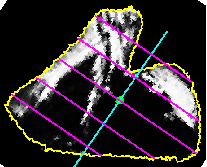}}
& 
\begin{itemize} [topsep = 0pt]
\item[] Region-wise Fat
  \begin{itemize}
  \item[] R1: 1.7 \% (Top), R2: 7.4 \%, R3: 3.5 \%, 
  \item[] R4: 2.1 \%, R5: 1.1 \%, R6: 0.6 \% 
  \end{itemize}
\item[] Total Fat : 15.5 \%
\item[] TCSA : 40 $mm^2$, FCSA : 34 $mm^2$
\end{itemize} \\
\hline

\raisebox{-\totalheight}
{\includegraphics[scale=0.9]{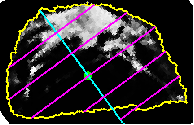}}
& 
\begin{itemize} [topsep = 0pt]
\item[] Region-wise Fat
  \begin{itemize}
  \item[] R1: 3.1 \% (Top), R2: 5.6 \%, R3: 3.1 \%,
  \item[] R4: 1.2 \%, R5: 0.3 \%, R6: 0.2 \%
  \end{itemize}
\item[] Total Fat : 13.2 \%
\item[] TCSA : 21 $mm^2$, FCSA : 18 $mm^2$
\end{itemize} \\
\hline
\end{tabular}
\end{table*}

\begin{table*}[!ht]
\centering
\caption{Region-wise fat quantification results in ES Muscles - Patient II}
\begin{tabular}{ l p{9cm} }
\hline
Segmented ES Muscles & Computations \\ 
\hline
\raisebox{-\totalheight}{\includegraphics[scale=0.8]{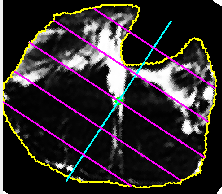}}
& 
\begin{itemize} [topsep = 0pt]
\item[] Region-wise Fat
  \begin{itemize}
  \item[] R1: 1.3 \% (Top), R2: 5.9 \%, R3: 4.7 \%,
  \item[] R4: 2.2 \%, R5: 1.0 \%, R6: 0.6 \%
  \end{itemize}
\item[] Total Fat : 14.8 \%
\item[] TCSA : 81 $mm^2$, FCSA : 69 $mm^2$
\end{itemize} \\
\hline

\raisebox{-\totalheight}
{\includegraphics[scale=0.8]{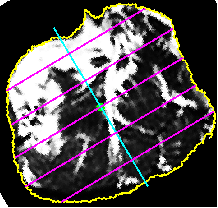}}
& 
\begin{itemize} [topsep = 0pt]
\item[] Region-wise Fat
  \begin{itemize}
  \item[] R1: 5.7 \% (Top), R2: 9.3 \%, R3: 7.2 \% ,
  \item[] R4: 5.1 \%, R5: 3.4 \%,  R6: 1.7 \% 
  \end{itemize}
\item[] Total Fat : 29.6 \%
\item[] TCSA : 79 $mm^2$, FCSA : 56 $mm^2$
\end{itemize} \\
\hline

\raisebox{-\totalheight}
{\includegraphics[scale=0.8]{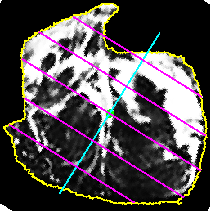}}
& 
\begin{itemize} [topsep = 0pt]
\item[] Region-wise Fat
  \begin{itemize}
  \item[] R1: 3.4 \% (Top), R2: 9.9 \%, R3: 9.9 \%
  \item[] R4: 5.5 \%, R5: 2.5 \%, R6: 0.9 \% 
  \end{itemize}
\item[] Total Fat : 29.8 \%
\item[] TCSA : 83 $mm^2$, FCSA : 58 $mm^2$
\end{itemize} \\
\hline

\raisebox{-\totalheight}
{\includegraphics[scale=0.8]{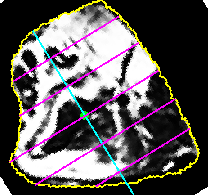}}
& 
\begin{itemize} [topsep = 0pt]
\item[] Region-wise Fat
  \begin{itemize}
  \item[] R1: 6.2 \% (Top), R2: 8.7 \%, R3: 10.0 \%,
  \item[] R4: 6.9 \%, R5: 6.0 \%, R6: 1.5 \% 
  \end{itemize}
\item[] Total Fat : 36.4 \%
\item[] TCSA : 52 $mm^2$, FCSA : 33 $mm^2$
\end{itemize} \\
\hline
\end{tabular}
\end{table*}

\subsection{Comparing fat percentage between pre-training and post-training sessions}
\begin{figure} [!ht]
\centering
\includegraphics[scale=0.5]{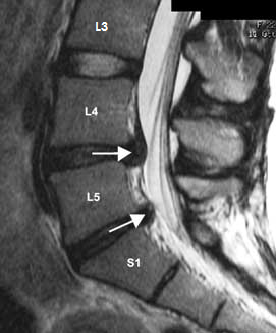}
\caption{Sagittal View of Lumbar Spine MRI showing spinal disc positions}
\label{fig:Disc}
\end{figure}

The MR images of two patients with lower back pain were acquired prior to  and after completion of the physical training sessions prescribed by the physicians. The MR images captured at the lumbar disc positions (L3-L4-L5-S1), as shown in Figure \ref{fig:Disc}, were considered for analysis using the tool. Even slight variations in the fat percentage of lumbar muscles between pre-training and post-training session could be easily identified using the tool, which were useful in determining the effectiveness of the training sessions. Tables 5 and 6 show the variations of fat percentage between pre-training and post-training session of patient I and patient II.

\vspace{-5mm}

\begin{table*}[!ht]
\centering
\caption{Comparison of fat percentage between pre-training and post-traning session - Patient I}
\begin{tabular}{l l c c c c c c c }
\hline
MRI Label & Training & Total Fat(\%)& R1(\%) & R2(\%) & R3(\%) & R4(\%) & R5(\%) & R6(\%)\\ 
\hline
L3L4(Left)
& Pre
& 6.0
& 1.3
& 1.2
& 1.0
& 1.0
& 0.8
& 0.8 \\

& Post
& 5.7
& 1.5
& 0.9
& 0.7
& 0.9
& 0.7
& 1.2 \\ 
\hline

L3L4(Right)
& Pre
& 7.3
& 1.5
& 1.3
& 1.1
& 1.4
& 1.3
& 0.9 \\

& Post
& 5.9
& 1.7
& 0.9
& 1.2
& 0.6
& 1.0
& 0.5 \\ 
\hline

L4L5(Left)
& Pre
& 7.8
& 2.2
& 1.2
& 1.0
& 0.8
& 1.5
& 1.3 \\

& Post
& 6.1
& 1.9
& 1.1
& 0.8
& 0.5
& 0.9
& 1.1 \\ 
\hline

L4L5(Right)
& Pre
& 9.0
& 1.7
& 2.3
& 1.4
& 1.4
& 1.7
& 0.8 \\

& Post
& 8.0
& 3.1
& 1.2
& 0.8
& 1.4
& 0.7
& 0.9 \\ 
\hline

L5S1(Left)
& Pre
& 15.4
& 5.8
& 5.3
& 2.5
& 1.7
& 0.8
& 0.5 \\

& Post
& 13.0
& 4.8
& 4.2
& 3.1
& 1.1
& 0.4
& 0.7 \\ 
\hline

L5S1(Right)
& Pre
& 18.3
& 5.2
& 6.3
& 4.1
& 2.4
& 1.0
& 0.8 \\

& Post
& 10.1
& 3.4
& 3.7
& 1.7
& 0.9
& 0.6
& 0.6 \\ 
\hline
\end{tabular}
\end{table*}

\begin{table*}[!ht]
\centering
\caption{Comparison of fat percentage between pre-training and post-traning session - Patient II}
\begin{tabular}{l l c c c c c c c }
\hline
MRI Label & Training & Total Fat(\%)& R1(\%) & R2(\%) & R3(\%) & R4(\%) & R5(\%) & R6(\%)\\ 
\hline
L3L4(Left)
& Pre
& 8.6
& 3.1
& 1.8
& 1.4
& 0.9
& 1.0
& 1.0 \\

& Post
& 6.7
& 2.2
& 1.3
& 1.1
& 0.5
& 0.9
& 1.1 \\ 
\hline

L3L4(Right)
& Pre
& 4.5
& 1.1
& 0.8
& 0.7
& 0.6
& 0.4
& 1.0 \\

& Post
& 2.9
& 0.7
& 0.5
& 0.5
& 0.3
& 0.4
& 0.7 \\ 
\hline

L4L5(Left)
& Pre
& 9.6
& 3.3
& 2.5
& 1.7
& 0.8
& 0.5
& 1.2 \\

& Post
& 9.5
& 3.2
& 2.8
& 1.8
& 0.9
& 0.5
& 0.8 \\ 
\hline

L4L5(Right)
& Pre
& 7.1
& 2.4
& 1.0
& 1.2
& 1.0
& 0.9
& 0.9 \\

& Post
& 6.6
& 2.4
& 0.9
& 0.9
& 0.8
& 0.6
& 1.3 \\ 
\hline

L5S1(Left)
& Pre
& 14.8
& 5.6
& 7.0
& 1.4
& 0.4
& 0.3
& 1.2 \\

& Post
& 13.4
& 7.1
& 4.9
& 1.2
& 0.4
& 0.3
& 0.7 \\ 
\hline

L5S1(Right)
& Pre
& 13.7
& 4.1
& 5.9
& 2.0
& 1.0
& 0.7
& 1.1 \\

& Post
& 10.6
& 5.2
& 3.3
& 1.1
& 0.7
& 0.4
& 0.8 \\ 
\hline
\end{tabular}
\end{table*}

\subsection{Graphical User Interface}
\begin{figure*}[!ht]
\centering
\includegraphics[scale=0.5] {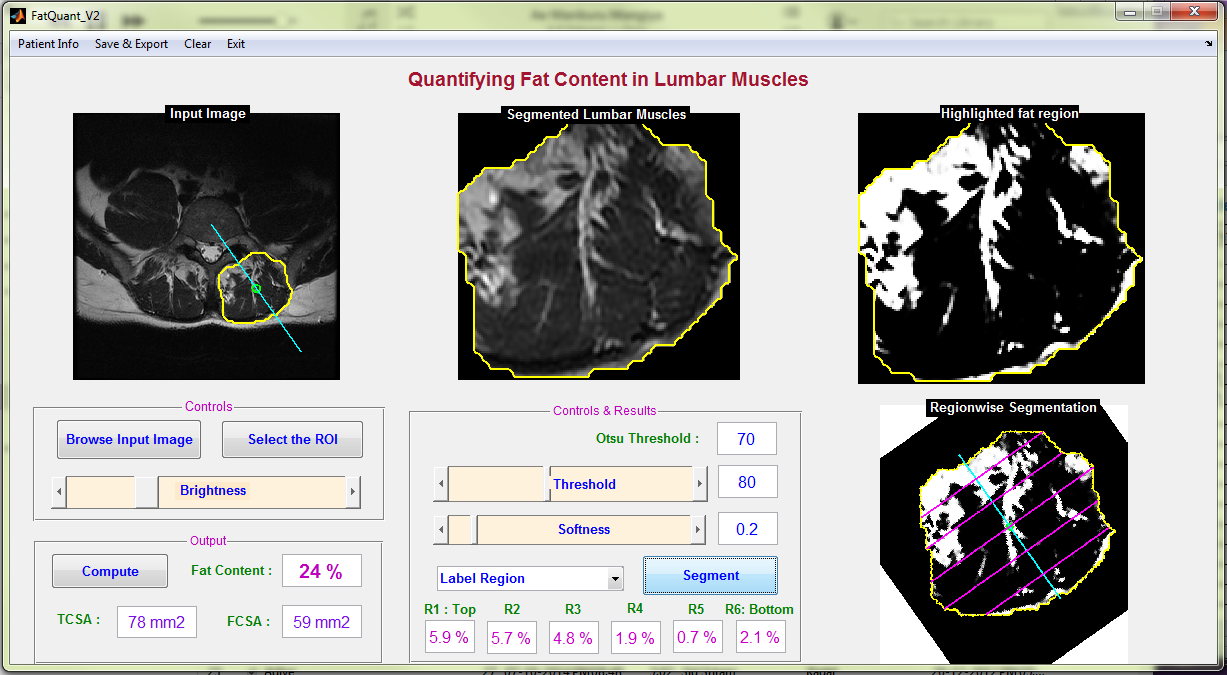}
\vspace{-3mm}
\caption{GUI screen shot: Quantifying fat content in ES Muscles.}
\label{fig:GUI Left ES}
\end{figure*}

A screen shot of the graphical user interface (GUI) captured while quantifying fat in erector spinae muscles is shown in Figure \ref{fig:GUI Left ES}.

\section{Conclusion}
We have proposed a method to quantify the cross-sectional area and distribution of fat in the MRI scans of lumbar muscles. We clearly show the interactive segmentation tool's advantage in precisely quantifying these measures over the commonly employed method of subjective evaluation (e.g. 5-point scale). This method will help future studies more accurately examine the relationship between fat infiltration and lower back pain; and if a relationship is evident, it may provide a greater insight into the rehabilitation process beyond reliance on a patient's reporting of pain. 

We have implemented the interactive segmentation of the erector spinae (ES) and the lumbar multifidus (LM) muscles using the livewire technique. The fat in the ES or LM muscles are discerned using two control parameters: softness and threshold via a sigmoid function. The threshold calculated using Otsu's method is taken as initial reference. 

We have also implemented a method to quantify the fat in a region wise manner with reference to the center of spinal column, which is automatically detected. This may be important as the distribution of fat and muscle relative to the axis of rotation produced by the whole muscle has functional implications for both the control of the spine and loading on the muscles. In addition, we have developed a graphical user interface (GUI) with interactive controls to perform the required computations, which can act as a supporting system to physicians.

Future work should focus on automatically segmenting the erector spinae (ES) and the Lumbar Multifidus (LM) muscles. The variations in the shape and size of ES and LM muscles in MRI slices of the same person are the key challenges for automatic segmentation. The accuracy in the automatic detection of the center of the spinal column can be improved by considering more training samples for the classifier model and/or using features based on pixel statistics, texture and transforms. 

\section*{Acknowledgement}

This publication has emanated from research conducted with the financial support of Science Foundation Ireland (SFI) under grant number SFI/12/RC/2289. 

\biboptions{sort&compress}
\bibliographystyle{model1-num-names}
\nocite{*}
\bibliography{Biblio}

\end{document}